\documentclass[sigconf]{acmart}
\AtBeginDocument{%
  }

\setcopyright{acmlicensed}
\copyrightyear{2025}
\acmYear{2025}
\acmDOI{}
\acmConference[KDD '25 Workshop on Structured Knowledge for LLMs]{KDD '25 Workshop on Structured Knowledge for LLMs}{August 04,
  2025}{Toronto, CA}
\acmISBN{}

\usepackage[utf8]{inputenc} 
\usepackage[T1]{fontenc}    
\usepackage{hyperref}       
\usepackage{url}            
\usepackage{booktabs}       
\usepackage{amsfonts}       
\usepackage{nicefrac}       
\usepackage{microtype}      
\usepackage{xcolor}         
\usepackage{graphicx, subcaption, multirow} 

\begin{document}

\title[Graph Representations for Flowchart QA]{A Graph-based Approach for Multi-Modal Question Answering from Flowcharts in Telecom Documents}

\author{Sumit Soman, H. G. Ranjani, Sujoy Roychowdhury, Venkata Dharma Surya Narayana Sastry, Akshat Jain, Pranav Gangrade, Ayaaz Khan}
\email{{sumit.soman, ranjani.h.g, sujoy.roychowdhury}@ericsson.com}
\orcid{1234-5678-9012}
\affiliation{%
  \institution{Ericsson R\&D}
  \city{Bangalore}
  \state{Karnataka}
  \country{India}
}








\renewcommand{\shortauthors}{Soman et al.}

\begin{abstract}
Question-Answering (QA) from technical documents often involves  questions whose answers are present in figures, such as flowcharts or flow diagrams. Text-based Retrieval Augmented Generation (RAG) systems may fail to answer such questions. We leverage graph  representations of flowcharts obtained from Visual large Language Models (VLMs) and incorporate them in a text-based RAG system to show that this approach can enable image retrieval for QA in the telecom domain. We present the end-to-end approach from processing technical documents, classifying image types, building graph representations, and incorporating them with the text embedding pipeline for efficient retrieval. We benchmark the same on a QA dataset created based on proprietary telecom product information documents. {\color{black}Results show that the graph representations obtained using a fine-tuned VLM model have lower edit distance with respect to the ground truth, which illustrate the robustness of these representations for flowchart images. Further, the approach for QA using these representations gives good retrieval performance using text-based embedding models, including a telecom-domain adapted one. Our approach also alleviates the need for a VLM in inference, which is an important cost benefit for deployed QA systems.}
\end{abstract}




\received{30 May 2025}

\maketitle

\section{Introduction}
\label{sec:intro}
With advances in Large Language Models (LLMs), Retrieval Augmented Generation (RAG) based  Question-Answering (QA) systems for technical domains like telecom have become prevalent \cite{roychowdhury2024evaluation, roychowdhury2024unlocking, soman2023observations}, which leverage domain-adapted models \cite{karapantelakis2024using, 10978393}. However, these works largely focus on text-based input and output modalities for RAG based QA task. A typical telecom engineer would require to refer to information present in {\color{black}Product Information Documents (referred to as PI Docs in this work)} and assimilate information from images (or figures), tables and/or equations, apart from text, to provide answers to user queries, or for troubleshooting scenarios. 

\begin{figure*}[t!]
    \centering
    \includegraphics[width=\linewidth]{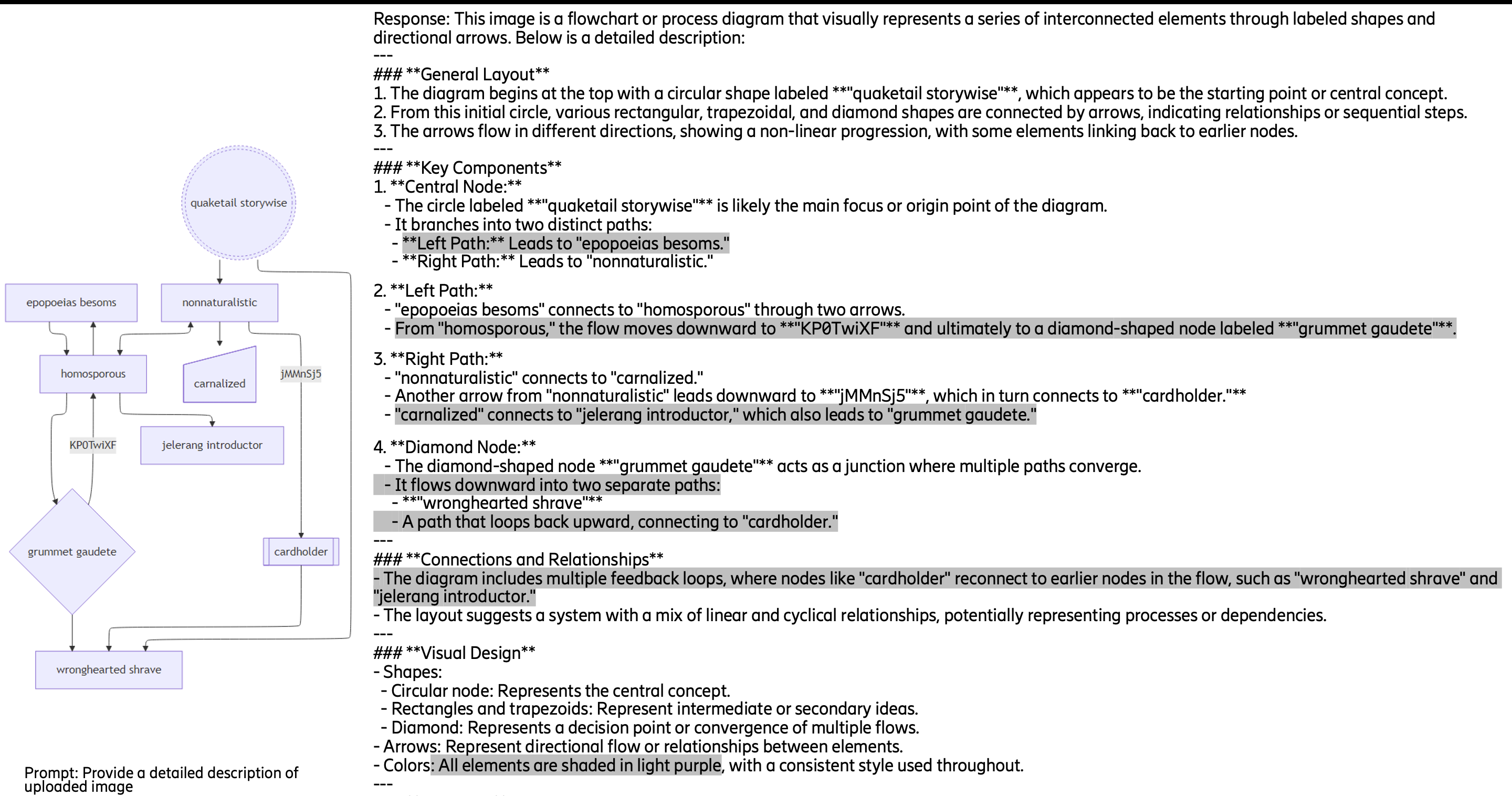}
    \caption{A sample description of synthetic flowchart from Flowlearn using GPT4 Vision model. The incorrect part of description are highlighted in gray. This illustrates the issues faced in using VLMs directly for technical QA involving flowcharts.}
    \label{fig:image_description}
\end{figure*}

\subsection{Background}
In the literature, multi-modality pertaining to images interspersed with text are typically addressed as Visual QA (VQA) tasks that use Visual Language Models (VLMs) \cite{keskar2025evaluating}. However, models trained on public datasets, which may include images of cat, dog, or background objects, are not useful for domains such as telecom. Typical images in telecom-domain documents pertain to categories such as screenshots of tools, flowcharts, block diagrams, sequence diagrams, icons, equations, schematic diagrams, among others, as can be seen in \cite{3gpp_release_18}. It is evident that images here often contain information primarily as text, and may also include icons, schematic representations, connections and dependencies among connected blocks or objects. The textual content is dominant and also domain-intensive.  

Popular RAG systems either support text modality alone or use multi-modal embeddings in vector database to support images, in addition to text in the input. However, the cost of using VLMs for embedding, and as the generator for RAG in an inference setup, is formidable for business needs due to the large size of VLMs \cite{li2024seed}. Approaches using knowledge graphs \cite{han2024retrieval}, as is common with textual data, require manual verification and hence are not suitable.

In this work, we focus on parsing, categorizing and processing images from proprietary technical documentation in the telecom domain.
We specifically focus on flowchart images which are hard to interpret. These flowcharts present a unique issue of having mostly text-based content and are also indicative of decision rules and conditions. These, in turn, are a very valuable source of information, especially in configuration and troubleshooting of telecom networks. However, summarizing a flowchart as text via VLMs might be tedious, prone to hallucinations, difficult to quantify accuracy, and potentially may lose out on various conditions {\color{black}(refer Figure \ref{fig:image_description})}. Converting flowcharts to graph structures has been addressed in the earlier works of \cite{pan2024flowlearn, singh2024flowvqa}. However, these works are limited to evaluating the conversion of flowcharts to graphs and evaluating the best representation of flowcharts (graphs or UML) for QA. The latter work, in fact, assumes that the correct flowchart is available for answering the question. Hence, there exists a gap in the study of retrieving the right flowchart or its representation. We build on some of these existing works to propose a solution which uses fine-tuned VLMs to convert flowcharts' images to graph representations, and benchmark it for retrieval for a typical QA task based on these flowcharts. 


\subsection{Problem Statement}

We propose the enhancement of text-only RAG with the following capabilities:
\begin{itemize}
    \item Categorize parsed images (from telecom documents - {\color{black}PI Docs}) into image categories identified as relevant for telecom domain QA. 
    \item Use a fine-tuned VLM to convert flowchart images to graph structures, using nodes connected through edges (unidirectional and/or bidirectional). The nodes and edges also have attributes associated. 
    \item Jointly represent flowchart based graph structures interspersed with text using LM domain-adapted embeddings.
    \item Utilize the graphical structures in RAG pipeline for improved coverage during retrieval. 
\end{itemize}


\begin{figure*}[ht]
    \centering
    \includegraphics[width=0.95\linewidth]{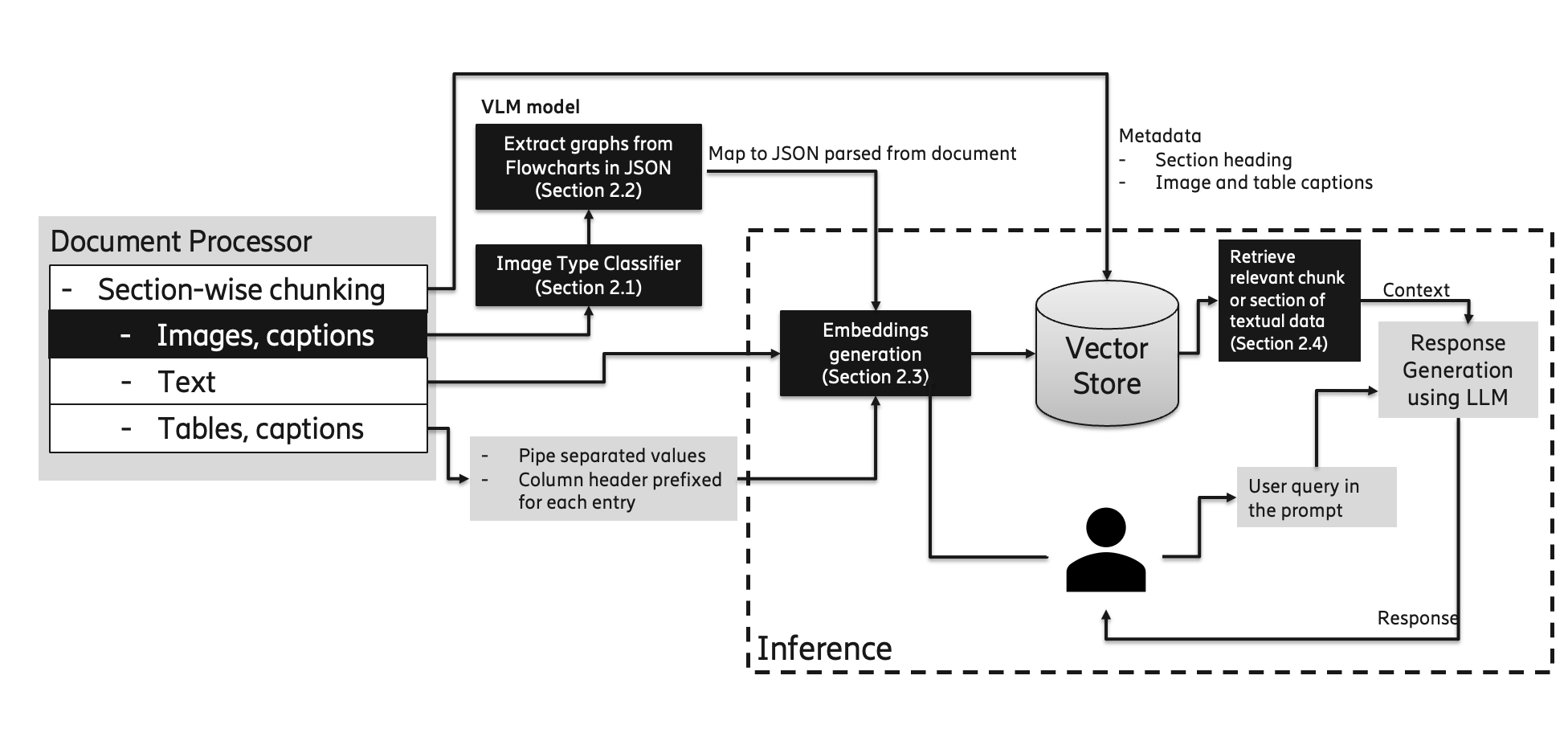}
    \caption{The end-to-end approach for multi-modal QA, our approach for images is indicated in black-shaded boxes.}
    \label{fig:overall_approach}
\end{figure*}
\subsection{Contributions}
The contributions of this work are as follows:
\begin{itemize}
    \item Automatically categorize images parsed from technical documentation using fine-tuned Document Image Transformer (DIT) model {\color{black}\cite{Lewis2006BuildingAT}}. 
    \item Convert flowchart images to graph structures using a fine-tuned VLM.
    \item Evaluate various chunking approaches to introduce these graphs in vector database for retrieval.
    \item Benchmark retrieval performance on QA dataset based on flowcharts.
\end{itemize}

{\color{black}The rest of the paper is organized as follows: Section \ref{sec:approach} details the proposed approach, with details of image classification (Section \ref{ssec:imageclass}), conversion of flowcharts to graphs (Section \ref{ssec:graphrep}), chunking and ingestion (Section \ref{ssec:chunk}) and evaluation (Section \ref{ssec:eval}). Experimental setup details are provided in Section \ref{sec:experiments} followed by the results and analysis in Section \ref{sec:results}}. Finally, conclusions and future work are presented in Section \ref{sec:conclusions}.

\section{Proposed Approach}
\label{sec:approach}
 As mentioned earlier, QA on flowcharts can be challenging using VLMs directly (since they may comprise information related to flow of information between nodes, decisions based on conditions or sequence of steps). Evaluation of these textual description of flowcharts can be challenging as there is no ground truth available and these descriptions can tend to be verbose based on complexity of flowcharts ({\color{black}refer Fig \ref{fig:image_description}}).
 
 Our proposed approach entails classification of images into various categories. Next, we consider only flowchart images. We use fine-tuned VLMs (trained using publicly available database) to convert these domain-specific flowchart images to graph representations. In the subsequent step, we use text-based chunking and obtain embeddings of these for retrieval. This section details the proposed approach and evaluation metrics. 
Fig. \ref{fig:overall_approach} depicts the proposed end-to-end approach for multi-modal QA. 
\subsection{Classification of images}
\label{ssec:imageclass}
Telecom documentation can have various categories of images. Typically, PI Docs processing involves parsing various formats (such as HTML, PDF) of documents and extracting the text, tables, equations and images from the paragraphs of various sections. Typically, the textual components are chunked (optimally) converted to embedding vector using domain-adapted embedding model {\color{black}\cite{soman2024observations}} and ingested into a vector database. 
In our approach, we aim to introduce only flowcharts using textual embedding for retrieval. To achieve this, we train a classifier to categorize the images parsed from PI Docs into various categories, \textit{viz.}, block diagrams, equations, flowcharts, graphs, hardware diagrams, icons (navigation, logos), schematic diagrams, screenshots, sequence diagrams. This helps categorize images for subsequent downstream QA task. In this work, we use the flowchart images for further processing and obtain their corresponding graph representation.

\subsection{Graph representation for flowcharts}
\label{ssec:graphrep}
We filter on the images identified as flowcharts from the classifier model discussed in Section \ref{ssec:imageclass}. On these, we propose to use a fine-tuned VLM to generate graph representations of the images.

\begin{figure*}[t]
    \centering
    \begin{subfigure}[t]{0.5\textwidth}
        \centering
        \includegraphics[width=0.5\linewidth]{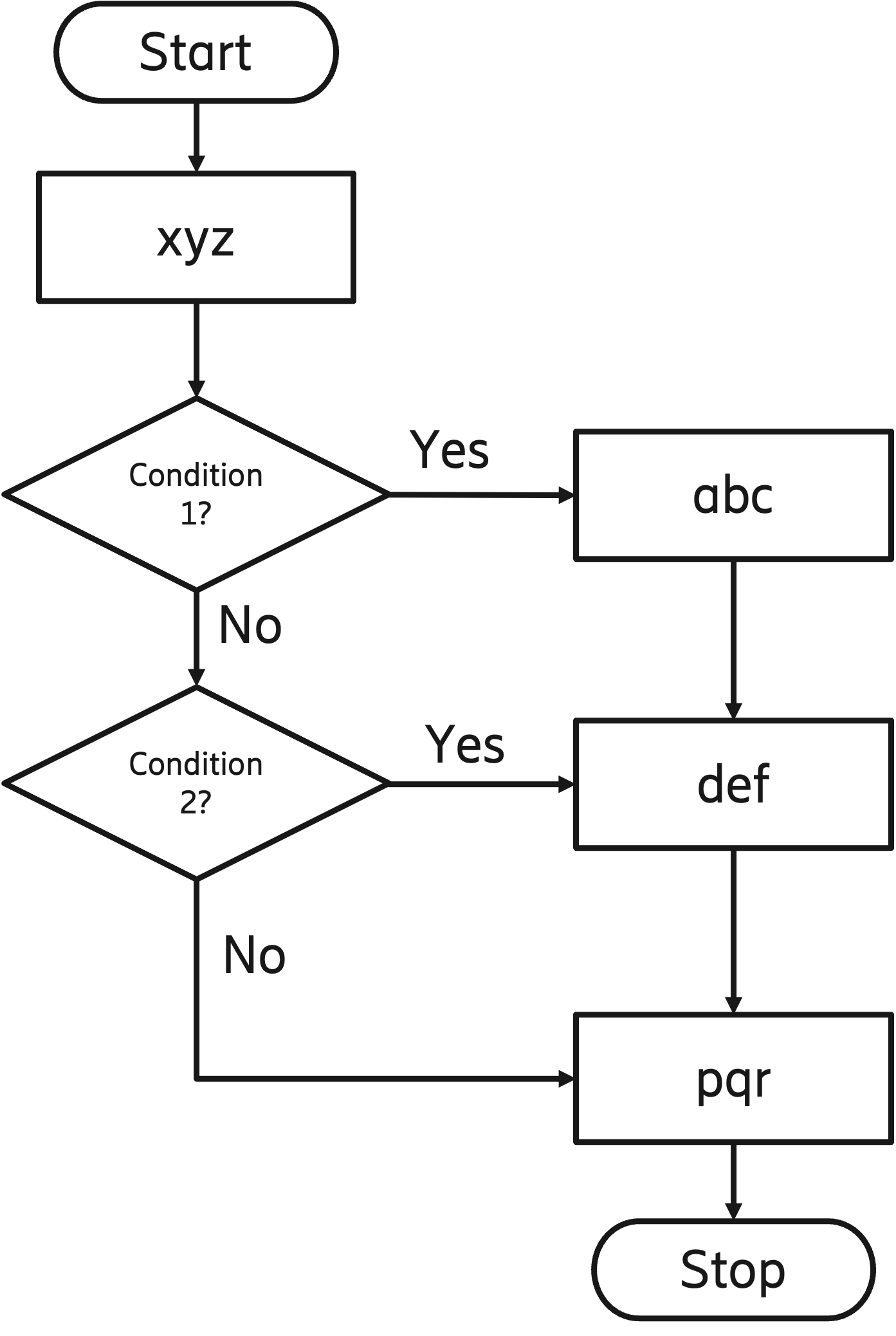}
        \caption{A sample image of a flowchart.}
        \label{fig:sample_flowchart}
    \end{subfigure}%
    ~ 
    \begin{subfigure}[t]{0.5\textwidth}
        \centering
        \includegraphics[width=0.95\linewidth]{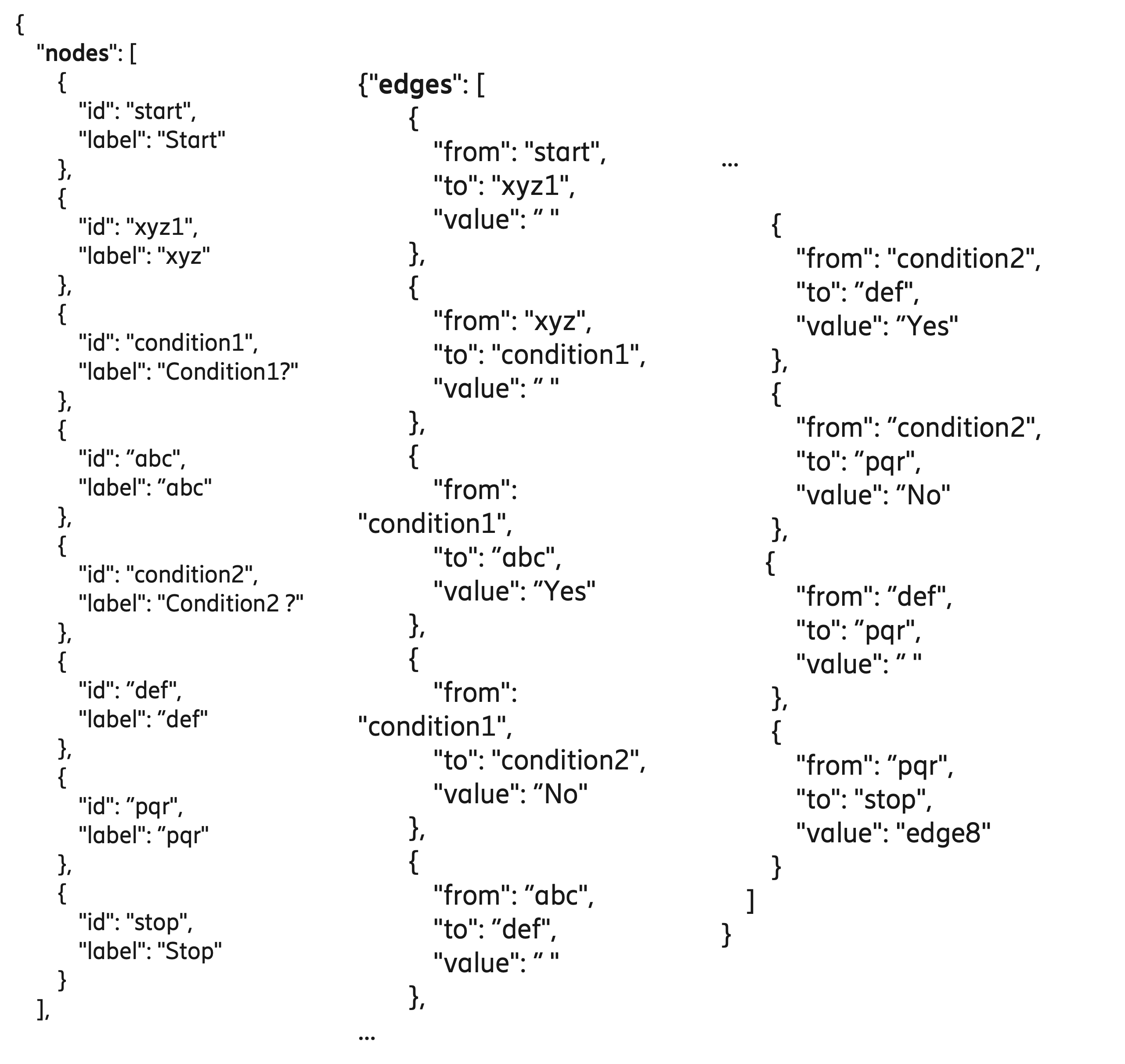}
        \caption{Graph representation in JSON (JavaScript Object Notation) with nodes and edges.}
        \label{fig:graph_json}
    \end{subfigure}
    \caption{A sample flowchart and its graph representation in JSON format.}
    \label{fig:flowchart_and_json}
\end{figure*}

A flowchart consists of a number of blocks and interconnecting links between them. We create a directed graph out of the flowchart - we represent each block as a node and capture the text within the block as a node attribute. Links between blocks are considered as edges, and any text on the link is considered as an edge attribute. Although flowcharts may have different shapes of blocks, we do not capture that in the node information. A sample representation of a flowchart and the corresponding graph (in JSON format) is shown in Figs. \ref{fig:sample_flowchart} and \ref{fig:graph_json}, respectively.

\subsubsection{Using a fine-tuned VLM}
In order to improve the graph representation generated by the open source VLM, we also fine-tuned 
a open-source VLM using the publicly available synthetic flowcharts of Flowlearn dataset \cite{pan2024flowlearn}. 

\begin{figure}
    \centering
    \includegraphics[width=0.95\linewidth]{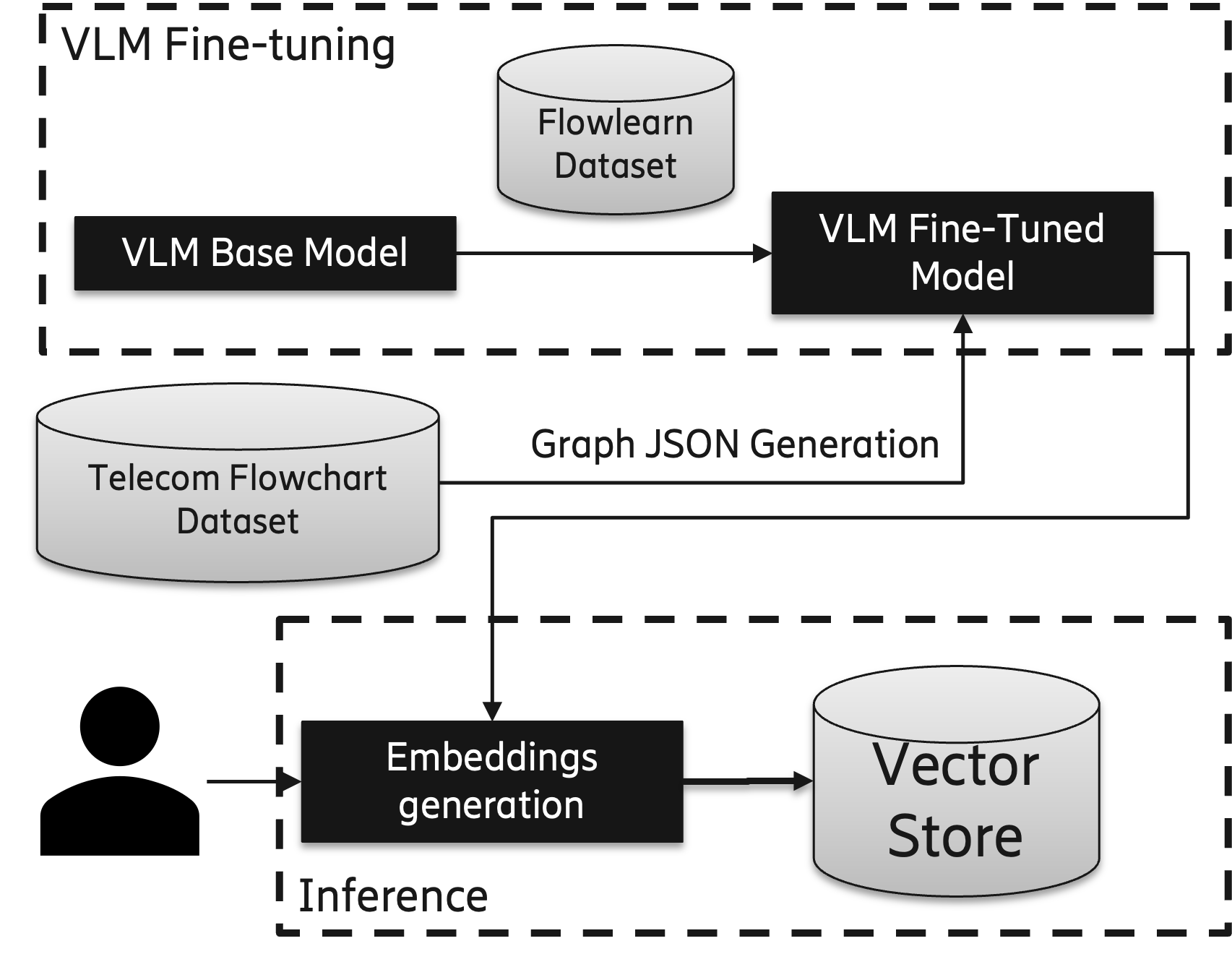}
    \caption{Fine-tuning the VLM with a publicly available dataset and using it on telecom data for generating graph representations.}
    \label{fig:vlm_finetuning}
\end{figure}

It may be noted here that the model was fine-tuned on a publicly available dataset of flowcharts and used to generate graphs for telecom-domain flowcharts, as shown in Fig. \ref{fig:vlm_finetuning}. 
Details of VLM considered and training dataset are given in Section \ref{sec:experiments}. 

\subsubsection{Evaluation metric for VLM}
The output of the VLM is a graph representations of the flowchart. Hence, we use Graph Edit Distance (GED) \cite{abu2015exact} as a measure of the performance of the model, apart from number of nodes and edges accurately detected. A lower value of GED indicates close similarity of the generated graph representation when compared with the ground-truth representation of that flowchart. We show in our results that we obtain lower GED with the fine-tuned VLMs.
 
 
\subsection{Chunking, ingestion into vector store and retrieval}
\label{ssec:chunk}

\begin{figure}[h]
    \centering
    \includegraphics[width=0.98\linewidth]{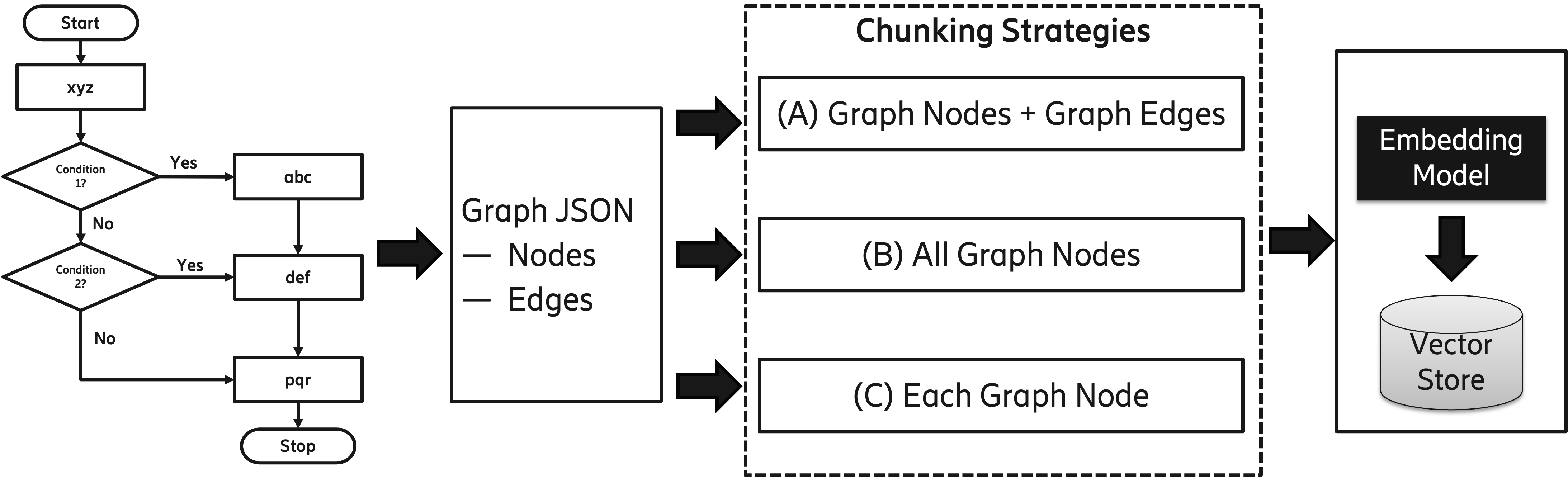}
    \caption{Chunking approaches for flowchart JSONs.}
    \label{fig:chunking_approaches}
\end{figure}

The graph structures obtained from the previous step (detailed in Section \ref{ssec:graphrep}) must be introduced into the vector store to ensure these  are included in the retriever stage of RAG. Hence, it is essential to be able to obtain embedding vectors for these graph structures. 
There have been studies which perform experiments to find optimal chunking of textual data and for tables to improve retrieval accuracy  \cite{soman2024observations}. Similarly, it is of importance to understand the optimal chunking mechanism for embedding graph structures. We consider the following options to generate embeddings, as shown in Fig. \ref{fig:chunking_approaches}:
\begin{itemize}
    \item \textbf{Each node as one chunk}: Embed \textit{each node's textual information} as a single embedding vector.
    \item \textbf{All nodes as one chunk}: Embed \textit{all the node textual information} as a single embedding vector.
    \item \textbf{Entire graph JSON as one chunk}: Embed the \textit{entire textual information} from graph JSON as a single embedding vector.
\end{itemize}

\subsection{Evaluation of Retrieval for RAG}
\label{ssec:eval}

We evaluate retrieval using two embedding models, the \textit{bge-large} \cite{bge_embedding} which is a publicly available embedding model and the \textit{TeleRoBERTa} \cite{holm2021bidirectional}, a telecom-domain adapted embedding model. For evaluating the retrieval, we compute top-$k$ accuracy for $k=\{1,3,5\}$. Following is the evaluation criteria adopted for the respective chunking approaches:
\begin{itemize}
\item When textual information of \textit{each node} is embedded as a vector, the retrieval is correct if \textit{any} of the nodes from the top-$k$ retrieved graph JSONs appears in the ground truth.
    \item When textual information of \textit{all the nodes} is embedded as a single vector, the retrieval is considered correct {\color{black}if \textit{all} of the nodes from the ground truth} appear in the top-$k$ retrieved graph JSONs.
    \item When the \textit{entire textual information} in the graph JSON is embedded as a single vector, the top-$k$ retrieval is considered correct one of the graph JSON among the $k$- retrieved graphs corresponds to the ground truth.
\end{itemize}

\section{Experiments}
\label{sec:experiments}
\subsection{Datasets}
\label{ssec:datasets}
For the dataset creation, we initially considered two sources -  (i) publicly available 3GPP (Rel 18) documents  (ii) proprietary PI Docs - both of which pertain to the telecom domain. 
\subsubsection{Image Sources}
From the 36 series of the publicly available 3GPP (Rel 18) documents \cite{3gpp_release_18}, we parse {\color{black}{6342}} images comprising of various categories such as  schematic diagrams, graphs, frequency plots, block diagrams, sequence diagrams. However, there are very few examples of flowcharts (less than 2\%) and many of these can be considered as a mix of flowchart as well as block diagram {\color{black}(refer Fig. \ref{fig:3gpp_flowchart} in Appendix \ref{apdx:3gpp_flowchart})}. Hence, we do not consider the flowcharts from this source, as it is not representative of the data we encounter during troubleshooting.

From the proprietary PI Docs, we have identified few documents to include in this dataset, extracted 1586 images of multiple types such as block diagrams, equations, flowcharts, graphs, hardware diagrams, icons (navigation, logos), schematic diagrams, screenshots, sequence diagrams, among others. Fig. \ref{fig:app_img_categories} shows sample images for the categories of images seen in PI Docs. Since this is a proprietary dataset, all content in images pertaining to PI Docs source have been obfuscated in this manuscript to retain confidentiality. Table {\ref{tab:image_categories}} shows the image categories considered and their statistics.
We have manually annotated this dataset to construct training and test sets with 1268 and 318 images respectively using stratified sampling across these image types. This dataset was used to fine-tune the classifier model that predicted the image type. 

\subsubsection{Image to graph representation}
For fine-tuning, we consider synthetic flowcharts component of Flowlearn dataset {\color{black}\cite{pan2024flowlearn}} for training VLM model. This consists of 10,000 flowchart images generated with \href{https://github.com/mermaid-js/mermaid}{\textit{Mermaid} tool}. The mermaid tool script is mapped to the required JSON format and is used for fine-tuning. However, we found that this dataset did not consider the following scenarios typically seen in flowcharts:
\begin{itemize}
    \item \textbf{Shapes of nodes} such as rectangles with semicircular ends, parallelograms, decision boxes (typically shaped as rhombus), connectors (shaped as circles and pentagons)
    \item \textbf{Edges,} including bi-directional edges, multiple arrow heads (small, medium, large), various edge connectors (solid, dotted, dashed), straight sharp edges
    \item \textbf{Node lines} - solid, dotted, dashed and sometimes, no outer lines
    \item \textbf{Edge attributes} with text associated
\end{itemize}
In order to deal with such scenarios, we synthetically created images using \href{https://github.com/mermaid-js/mermaid}{\textit{Mermaid} tool}, using existing Flowlearn dataset as the starting point. These synthetically generated images were augmented with existing images of Flowlearn dataset. The augmented images for training (fine-tuning) and testing are kept separate.  

\subsubsection{QA Dataset for Retrieval}   
For testing retrieval accuracy, we consider 105 flowchart images from PI Docs which have been associated with ground truth graph structures. With inputs from Subject Matter Experts (SMEs), we carefully curate a set of 502 QA pairs from these images, with a mode of $\sim5$ questions for each image. Each QA pairs is tagged as `Decision related' (D), `Edge related' (E) and `Node related' (N) 
- based on how to arrive at the answer from the question. Details of the QA dataset are listed in {\color{black}Table \ref{tab:GQA}} 
for different hops. The categories listed are based on ability of retriever to identify the correct chunk of data which contains the graphical structure from the flowchart image. 

\begin{table}[h]
\begin{tabular}{|ll|}
\hline
 Category& \# of QA\\\hline\hline
  Decision related   & 359 \\
   Node related  &487\\
 Edge related    & 479 \\
 \hline
\end{tabular}
\caption{Distribution of QA categories considered}
\label{tab:GQA}
\end{table}


\subsection{Experimental Setup}
\label{ssec:exptal_setup}
\subsubsection{Image Category Classifier}
We use the manually annotated PI Docs image dataset (1586 images) described in Section \ref{ssec:datasets} to train (fine-tune) the image classification model. This dataset is split into train and test dataset in the ratio {\color{black}80-20 split} (1268-318 train-test split) using stratified sampling. 
We fine-tune the ``\textit{microsoft/dit-base}'' model \cite{Lewis2006BuildingAT} with batch size of $16$ for this dataset. 

While it is possible to explore models other than DIT, we note that image classification is not the primary focus of this work and this model performance can be considered as a baseline for further improvements. 
\subsubsection{VLM for graph representation}
The top-performing open-source VLMs \footnote{Since PI Docs are proprietary in nature, it is preferable to use open-source VLMs to avoid data-sharing outside the organization} available at the time of conducting our experiments were \textit{Qwen2-VL} \cite{wang2024qwen2} and \textit{Llava 1.5} {\color{black}\cite{ liu2024llavanext}}. We considered \textit{Qwen2-VL} for fine-tuning due to better performance on few samples. 
The prompt used for generating the graph representation of a flowchart is:

\begin{quote}
    \textit{``I have uploaded an image of a flowchart and here is its ground truth JSON representation, image\_json =\{\}.}
\textit{Now generate JSON for the next image, from and to should be the node IDs. In the edges section, make sure that the edge value is present. If there are multiple identical nodes, create different IDs for them and their edges accordingly.''}
\end{quote}

The synthetic flowcharts from Flowlearn dataset \cite{pan2024flowlearn} is considered for fine-tuned VLM. It has {\color{black}10,000} images, split as {\color{black}64-16-20\% for train, validation and test respectively}. We augment this training set with synthetic data for improved coverage of various nodes and edges (details described in \ref{ssec:datasets}). The fine-tuning was performed for few choices of parameters $R$ and $\alpha$, and the best fine-tuned model was used in the pipeline.
\subsubsection{Retrieval with chunking approaches}
As detailed in Section \ref{ssec:eval}, we evaluate various chunking approaches via retrieval accuracies. Three chunking approaches are proposed and evaluated for top-\textit{k} accuracies. The chunking approaches are evaluated in two scenarios: (i) Embedding vectors of only graph-structures are considered for retrieval {\color{black}(ii) Embeddings of graph-structure and accompanying text are considered for retrieval}. This results in 6 variations of retrieval for each model. We consider two embedding models - \textit{bge-large} and TeleRoBERTa (domain-adapted) \cite{bge_embedding,holm2021bidirectional} and hence will have {\color{black}12} sets of retrieval results. 
\section{Results and Analysis}
\label{sec:results}
In this section, we tabulate and detail results of the experiments listed in experimental setup section. 
\subsection{Image Category Classifier}
We consider dataset of 1586 images from the proprietary dataset and labeled them manually to create  training and test datasets. The distribution of number of images in the respective categories is shown in Table \ref{tab:image_categories}. We fine-tuned the ``\textit{microsoft/dit-base}'' model \cite{Lewis2006BuildingAT} with batch size of $16$ for this dataset.

\begin{table}[h]
\centering
\begin{tabular}{|l|l|l|l|l|}
\hline
S. No. & Class             & \#Train & \#Test & Total \\ \hline
1. & Block Diagram     & 123     & 39     & 162   \\ 
2. & Equation          & 154     & 42     & 196   \\ 
3. & Flowchart         & 171     & 38     & 209   \\ 
4. & Graph             & 41      & 9      & 50    \\ 
5. & Hardware          & 7       & 3      & 10    \\ 
6. & Icon              & 15      & 3      & 18    \\ 
7. & Others            & 131     & 43     & 174   \\ 
8. & Schematic Diagram & 168     & 29     & 197   \\ 
9. & Screenshot        & 388     & 102    & 490   \\ 
10. & Sequence Diagram  & 70      & 10     & 80    \\ \hline
 & Total             & 1268    & 318    & 1586  \\ \hline
\end{tabular}
\caption{Image categories for telecom dataset.}
\label{tab:image_categories}
\end{table}

\begin{figure*}[h]
    \centering
    \includegraphics[width=0.80\textwidth]{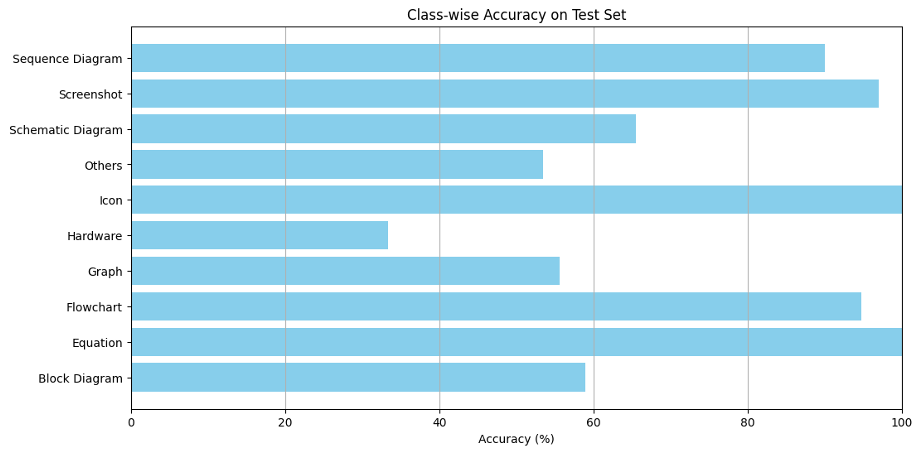}
    \caption{Performance of the classifier on the test set for various image categories from PI docs.}
    \label{fig:classifier_performance}
\end{figure*}

The fine-tuned model was evaluated on the test set for predicting the image categories. The performance of the model on various categories of images is shown in Fig. \ref{fig:classifier_performance}. We observe that the accuracy of prediction is high for images categories like icons and equations, while it is above 80\% for sequence diagrams, screenshots and flowcharts.

\begin{table*}[t]
\scalebox{1.0}{%
\begin{tabular}{|l|l|l|l|l|l|}
\hline
Model                               & \begin{tabular}[c]{@{}l@{}}Avg. \#Nodes\\ (Ground Truth)\end{tabular} & \begin{tabular}[c]{@{}l@{}}Avg. \#Edges\\ (Ground Truth)\end{tabular} & \begin{tabular}[c]{@{}l@{}} Avg. \#Nodes \\ Detected \end{tabular} & \begin{tabular}[c]{@{}l@{}} Avg. \#Edges \\ Detected \end{tabular} & \begin{tabular}[c]{@{}l@{}} Avg. Graph \\ Edit Distance (GED)  \end{tabular}          \\ \hline

\multicolumn{6}{|c|}{Flowcharts from test set of Flowlearn dataset}\\
\hline
Qwen2-VL Base                       & 6.36                                                               & 6.82                                                               & 6.55                 & 8.57                  & {\color[HTML]{333333} 10.21}         \\ 
Qwen2-VL FT - Lora R=8, Alpha=16    & 6.36                                                               & 6.82                                                               & 6.3                  & 6.23                  & {\color[HTML]{333333} 4.24}          \\ 
Qwen2-VL FT - Lora R=32, Alpha=32   & 6.36                                                               & 6.82                                                               & 6.2                  & 6.53                  & {\color[HTML]{333333} 5.09}          \\ 
Qwen2-VL FT - Lora R=128, Alpha=128 & 6.36                                                               & 6.82                                                               & 6.34                 & 6.43                  & {\color[HTML]{333333} 4.43}          \\ 
Qwen2-VL FT - Lora R=256, Alpha=256 & 6.36                                                               & 6.82                                                               & 6.35                 & 6.7                   & {\color[HTML]{333333} 3.82}          \\ 
Qwen2-VL FT - Lora R=512, Alpha=512 & 6.36                                                               & 6.82                                                               & 6.36                 & 6.42                  & {\color[HTML]{333333} \textbf{2.74}} \\
\hline
\multicolumn{6}{|c|}{Flowcharts from PI Docs}\\\hline
Qwen2-VL FT - Lora R=512, Alpha=512 & 12.54                                                               & 11.77                                                               & 12.32                 & 11.11                 & {\color[HTML]{333333} \textbf{3.14}} \\
\hline 
\end{tabular}%
}
\caption{Graph Metrics for fine-tuned VLM with various parameter settings reported for {\color{black} test set of Flowlearn dataset} and flowcharts from PI Docs.}
\label{tab:vlm_finetuning}
\end{table*}

\begin{table*}[!ht]
\begin{tabular}{|l|lll|lll|}
\hline
\multicolumn{1}{|c|}{Embedding Model}            & \multicolumn{3}{c|}{bge-large}                                                 & \multicolumn{3}{c|}{TeleRoBERTa}                                                                \\ \hline
\multicolumn{1}{|c|}{Chunking Approach} & \multicolumn{1}{l|}{Top-1}   & \multicolumn{1}{l|}{Top-3}   & Top-5            & \multicolumn{1}{l|}{Top-1}            & \multicolumn{1}{l|}{Top-3}            & Top-5   \\ \hline
\multicolumn{7}{|c|}{Embeddings of graph structures only for retrieval}\\\hline
Each node as one chunk                               & \multicolumn{1}{l|}{\textbf{56.17}\%} & \multicolumn{1}{l|}{65.33\%} & 66.93\%          & \multicolumn{1}{l|}{\textbf{57.17\%}} & \multicolumn{1}{l|}{62.74\%}          & 65.93\% \\ 
All the nodes as one chunk                       & \multicolumn{1}{l|}{50.29\%} & \multicolumn{1}{l|}{68.12\%} & 75.23\%          & \multicolumn{1}{l|}{49.80\%}          & \multicolumn{1}{l|}{\textbf{71.91\%}} & \textbf{76.89}\% \\ 
Entire graph JSON as one chunk                          & \multicolumn{1}{l|}{53.19\%} & \multicolumn{1}{l|}{\textbf{71.12}\%} & \textbf{78.29\%} & \multicolumn{1}{l|}{49.06\%}          & \multicolumn{1}{l|}{71.71\%}          & 76.69\% \\ \hline

\multicolumn{7}{|c|}{Embeddings of graph structures interspersed with text for retrieval}\\\hline
Each node as one chunk      &\multicolumn{1}{l|}{\textbf{41.05}}&\multicolumn{1}{l|}{\textbf{55.53}}&\multicolumn{1}{l|}{\textbf{59.76}}&\multicolumn{1}{l|}{\textbf{32.42}}&\multicolumn{1}{l|}{\textbf{44.08}}&\textbf{48.86}\\
All the nodes as one chunk  &\multicolumn{1}{l|}{38.84}&\multicolumn{1}{l|}{43.63}&\multicolumn{1}{l|}{43.82}&\multicolumn{1}{l|}{30.92}&\multicolumn{1}{l|}{38.33}&42.29\\
Entire graph JSON as one chunk &\multicolumn{1}{l|}{30.08}&\multicolumn{1}{l|}{31.47}&\multicolumn{1}{l|}{32.03}&\multicolumn{1}{l|}{24.27}&\multicolumn{1}{l|}{28.15}&39.83\\\hline
\end{tabular}
\caption{Retriever performance of chunking approaches for the embedding models, best top-$k$ values are indicated in bold {\color{black}for (i) only embeddings of graph structures considered for retrieval, and (ii) with embeddings of graph structures and text considered for retrieval.}}

\label{tab:retriever_results}
\end{table*}

\begin{table*}[h]
\begin{tabular}{|l|lllllllll|}
\hline

\multicolumn{1}{|c|}{Chunking Approach}          & \multicolumn{3}{c|}{Top-1}                                                                          & \multicolumn{3}{c|}{Top-3}                                                                          & \multicolumn{3}{c|}{Top-5}                                                                          \\ \hline
\multicolumn{1}{|c|}{Question Category}          & \multicolumn{1}{c|}{$D$}          & \multicolumn{1}{c|}{$N$}          & \multicolumn{1}{c|}{$E$}          & \multicolumn{1}{c|}{$D$}          & \multicolumn{1}{c|}{$N$}          & \multicolumn{1}{c|}{$E$}          & \multicolumn{1}{c|}{$D$}          & \multicolumn{1}{c|}{$N$}          & \multicolumn{1}{c|}{$E$}          \\ \hline
\multicolumn{10}{|c|}{bge-large}                                                                                                                          \\ \hline
Each node as one chunk                               & \multicolumn{1}{l|}{46.11}      & \multicolumn{1}{l|}{\textbf{59.20}}      & \multicolumn{1}{l|}{40.67}      & \multicolumn{1}{l|}{63.72}      & \multicolumn{1}{l|}{\textbf{77.53}}      & \multicolumn{1}{l|}{60.03}      & \multicolumn{1}{l|}{66.22}      & \multicolumn{1}{l|}{\textbf{81.43}}      & 63.61                           \\ 
All the nodes as one chunk                       & \multicolumn{1}{l|}{46.96}      & \multicolumn{1}{l|}{52.44}      & \multicolumn{1}{l|}{47.88}      & \multicolumn{1}{l|}{65.12}      & \multicolumn{1}{l|}{68.27}      & \multicolumn{1}{l|}{65.61}      & \multicolumn{1}{l|}{70.33}      & \multicolumn{1}{l|}{78.19}      & 66.42                           \\ 
Entire graph JSON as one chunk                          & \multicolumn{1}{l|}{\textbf{47.35}}      & \multicolumn{1}{l|}{52.57}      & \multicolumn{1}{l|}{\textbf{51.98}}      & \multicolumn{1}{l|}{\textbf{67.97}}      & \multicolumn{1}{l|}{70.43}      & \multicolumn{1}{l|}{\textbf{70.15}}      & \multicolumn{1}{l|}{\textbf{71.49}}      & \multicolumn{1}{l|}{77.41}      & \textbf{77.45}                           \\ \hline
\multicolumn{10}{|c|}{TeleRoBERTa}                                                                                                                                                                                                                                                                      \\ \hline
Each node as one chunk                               & \multicolumn{1}{l|}{43.21}      & \multicolumn{1}{l|}{\textbf{56.71}}      & \multicolumn{1}{l|}{37.02}      & \multicolumn{1}{l|}{58.12}      & \multicolumn{1}{l|}{\textbf{73.72}}      & \multicolumn{1}{l|}{55.73}      & \multicolumn{1}{l|}{63.32}      & \multicolumn{1}{l|}{\textbf{78.55}}      & 59.61                           \\ 
All the nodes as one chunk                       & \multicolumn{1}{l|}{\textbf{47.13}}      & \multicolumn{1}{l|}{50.29}      & \multicolumn{1}{l|}{49.77}      & \multicolumn{1}{l|}{61.36}      & \multicolumn{1}{l|}{65.22}      & \multicolumn{1}{l|}{64.44}      & \multicolumn{1}{l|}{67.80}      & \multicolumn{1}{l|}{73.41}      & 65.16                           \\ 
Entire graph JSON as one chunk                          & \multicolumn{1}{l|}{46.75}      & \multicolumn{1}{l|}{49.20}      & \multicolumn{1}{l|}{\textbf{51.87}}      & \multicolumn{1}{l|}{\textbf{65.64}}      & \multicolumn{1}{l|}{66.09}      & \multicolumn{1}{l|}{\textbf{69.08}}      & \multicolumn{1}{l|}{\textbf{71.22}}      & \multicolumn{1}{l|}{72.34}      & \textbf{70.77}                           \\ \hline
\end{tabular}
\caption{$D$, $N$, $E$ indicative of Decision based, Node based and Edge based QA. Retriever performance of chunking approaches for the two embedding models. Best top-$k$ values are indicated in bold for various chunking approaches.}
\label{tab:category_accuracy}
\end{table*}

\begin{figure}[h]
    \centering
    \includegraphics[width=0.99\linewidth]{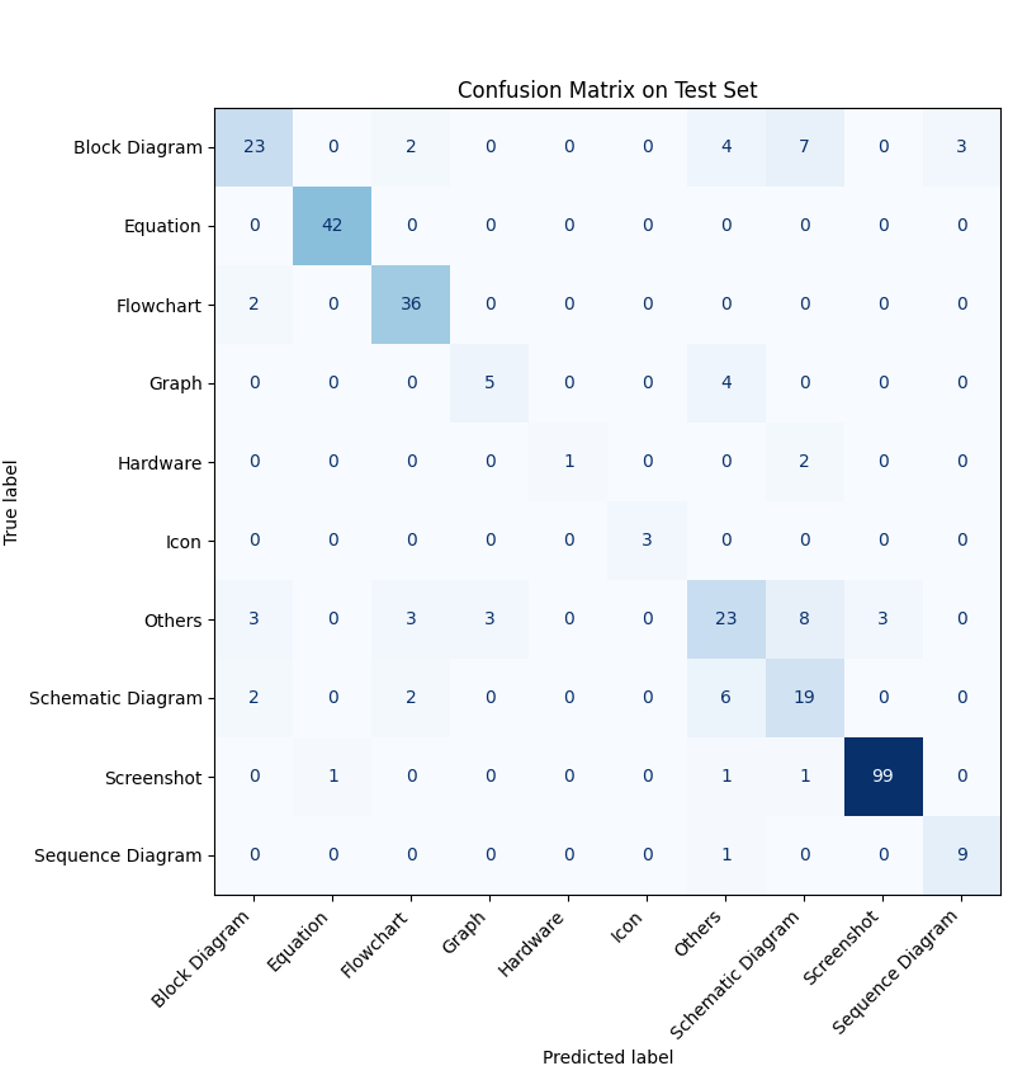}
    \caption{Confusion matrix on the test set for categorization of images from the PI dataset.}
    \label{fig:classifier_confusion_matrix}
\end{figure}

We also show the confusion matrix for the test set indicating the correct and incorrect classification of images in the test set in Fig. \ref{fig:classifier_confusion_matrix}. We observe that most of the flowchart images are classified correctly, while some are misclassified as block diagram, schematic diagram or others. This is expected since these images can be similar and belong to multiple categories. However, since most of the flowchart images are categorized correctly, we use this classifier in the pipeline. The performance can potentially be improved as more annotated images are available for fine-tuning.

\subsection{VLM for graph representation}

The results for fine-tuning of the \textit{Qwen2-VL}  \cite{wang2024qwen2} VLM are shown in Table \ref{tab:vlm_finetuning}, for various parameter values. The columns indicate the average number of nodes and edges in the ground truth graph JSON representations, the number of nodes after the transformation operations and the number of edges detected for the model outputs. The GED metric is shown in the last column. We obtain the lowest GED of $2.74$ using both Lora R and $\alpha$ as 512 for the finetuned \textit{Qwen2-VL} model, which is a significant improvement over the base model which has a GED of $10.21$ on the test set.

{\color{black} Similar performance evaluation of flowcharts from PI Docs is reported using the best performing fine-tuned model in Table \ref{tab:vlm_finetuning}. We observe that the average number of nodes and edges in the flowcharts from PI Docs is almost twice that of those seen in Flowchart dataset. Results show that GED on this unseen flowchart data is quite low ($3.14$)}. 

\subsection{Retrieval with chunking approaches}

Table \ref{tab:retriever_results} shows the retriever performance on top-$k$ accuracy for the chunking approaches using a publicly available base model (\textit{bge-large}) and domain-adapted (\textit{TeleRoBERTa}) embedding models. We highlight here that the \textit{TeleRoBERTa} model has been domain-adapted on publicly available telecom data (3GPP), and does not include any images related information during its training phase. Hence, there is no data contamination for the data considered and evaluated in this work. The objective here is to compare it\footnote{\textit{TeleRoberta} model is much smaller than \textit{bge-large} model in terms of parameter size.} with a publicly available embedding model.  We observe that the best top-$k$ retrieval results for $k=\{1,3\}$ are obtained when using for \textit{TeleRoBERTa} (57.17\% and 71.91\% respectively). Further, using each node as one chunk (embedding vector) gives better results for top-1 (57.17\%) for \textit{TeleRoBERTa}. 

We also observe that retrieval accuracy reduces in the scenario when embeddings of graph structures are interspersed with text in the vector store.  This is expected, as in a typical scenario, the retrieved top-$k$ embeddings are no longer limited to only graph structures. 

Table \ref{tab:category_accuracy} shows the top-$k$ retrieval accuracy for the embedding models for the various QA categories. Across the chunking strategies, higher performance is most commonly seen when using the entire graph JSON as one chunk, for both models. Better performance is obtained for node-related questions, followed by decision-related and finally, edge-related questions.

\section{Conclusions and Future Work}
\label{sec:conclusions}
{\color{black}
In this work, we have considered an approach to introduce flowchart images with dominant textual content into retrieval (for RAG) using textual embeddings. We first categorize images present in the domain dataset using a fine-tuned DIT model. We observe that the accuracy of flowchart category of images is sufficiently high. Next, the flowchart images are converted to graph structures using a fine-tuned VLM. Here, we show that the fine-tuned VLM has lower GED for the flowchart graph representations. These graph structures are then embedded into text-based vector store and benchmarked for retrieval accuracy on a QA dataset based on these graphs. We observe that embedding the whole graph as one vector shows higher accuracy. This is because the chunk of data embedded includes all the information related to the nodes and edges of the graph. This is also shown to have better performance when the QA category is node-related. 

Future work includes 
evaluation of generator output when JSON structures are passed as context to textual generator component of RAG. Analyzing cases involving errors in graph generation and retrieval, and performance with interspersed (document) text would be of interest. Additionally, extending the capabilities to other types of diagrams like UML Sequence diagrams which have a semantic structure are areas of potential future research for the community.}

\bibliographystyle{ACM-Reference-Format}
\bibliography{sample-base}




\appendix

\section{Image Categories in Telecom Dataset}

\begin{figure*}[t!]
    \centering
    \begin{subfigure}[t]{0.5\textwidth}
        \centering
        \includegraphics[width=0.95\columnwidth]{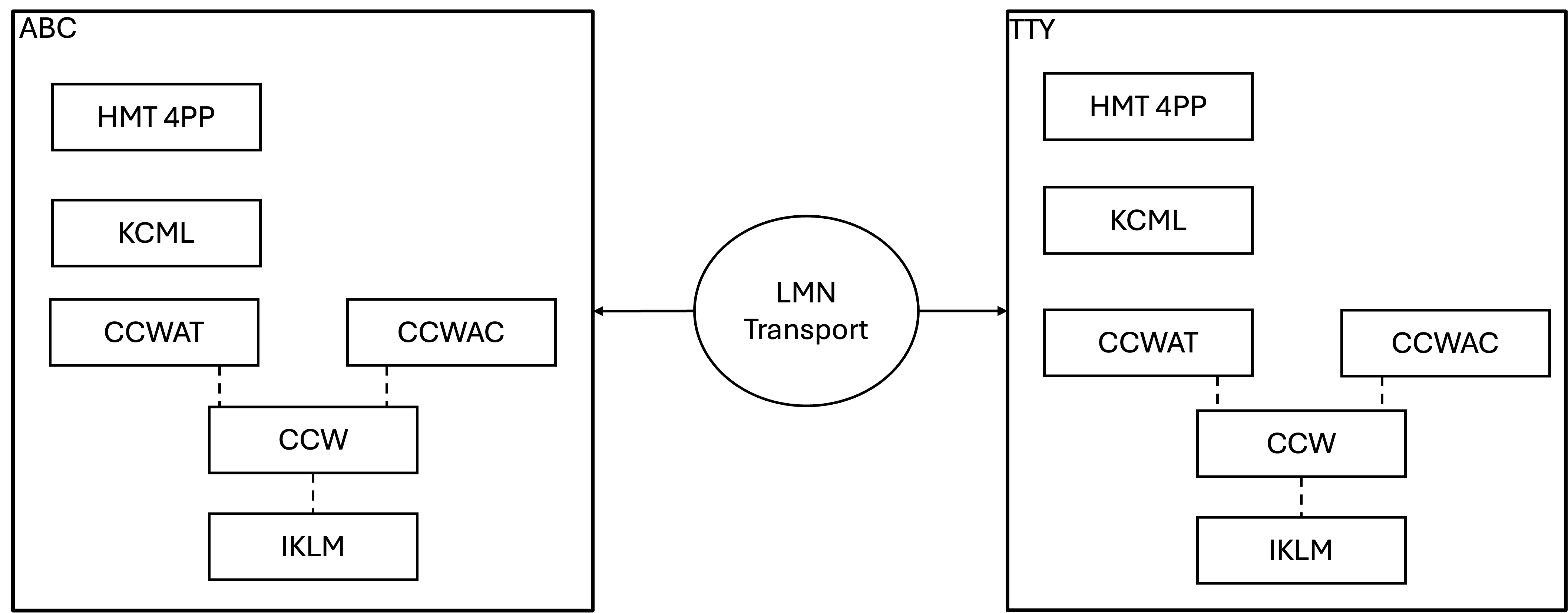}
        \caption{Block Diagram}
    \end{subfigure}%
    ~ 
    \begin{subfigure}[t]{0.5\textwidth}
        \centering
        \includegraphics[width=0.9\columnwidth]{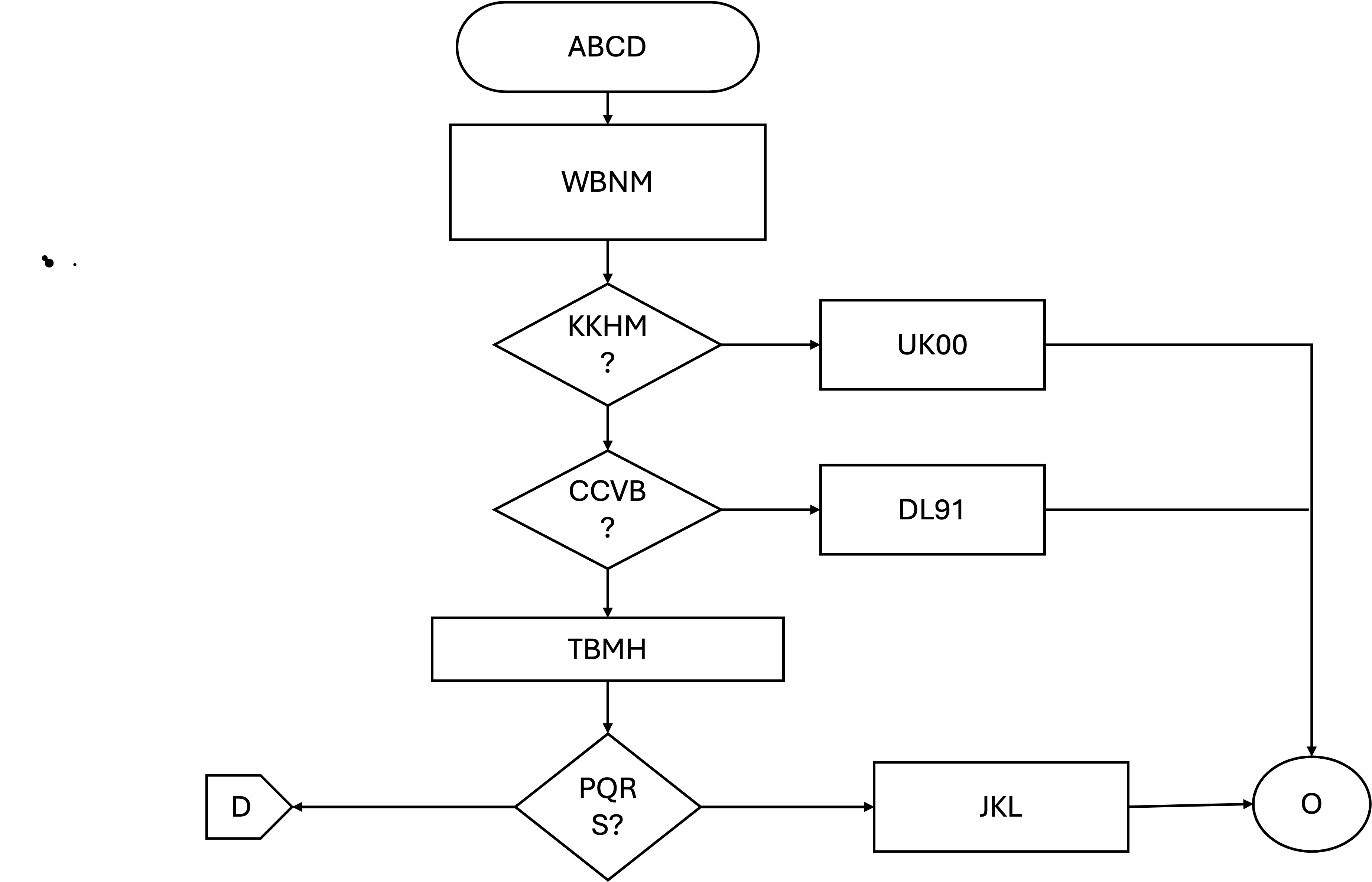}
        \caption{Flowchart}
    \end{subfigure}

        \begin{subfigure}[t]{0.32\textwidth}
        \centering
        \includegraphics[width=0.8\columnwidth]{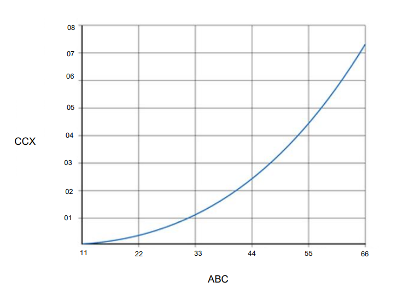}
        \caption{Graphs}
    \end{subfigure}%
    ~ 
    \begin{subfigure}[t]{0.32\textwidth}
        \centering
        \includegraphics[width=0.65\columnwidth]{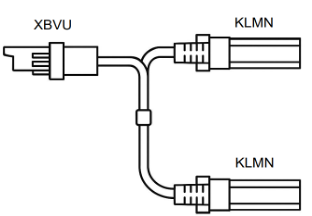}
        \caption{Hardware Diagrams}
    \end{subfigure}
    ~
    \begin{subfigure}[t]{0.32\textwidth}
        \centering
        \includegraphics[width=0.35\columnwidth]{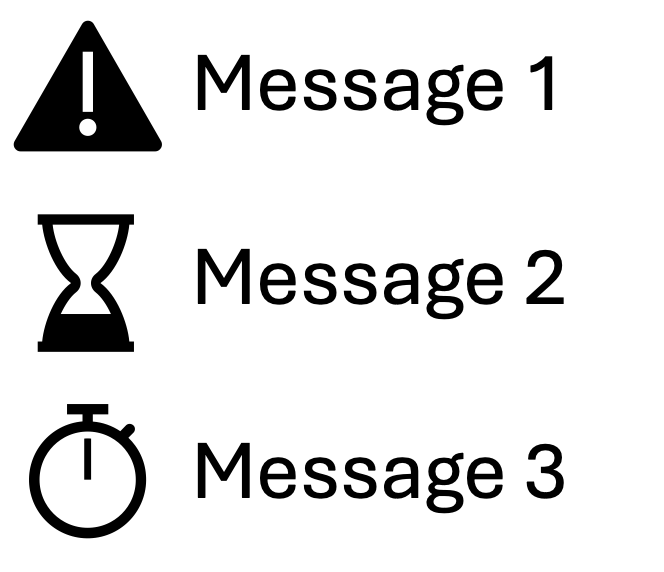}
        \caption{Icons}
    \end{subfigure}%

    \begin{subfigure}[t]{0.32\textwidth}
        \centering
        \includegraphics[width=0.65\columnwidth]{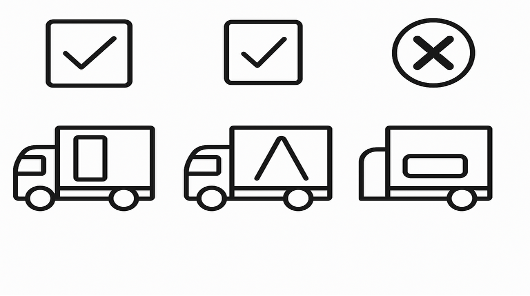}
        \caption{Instruction Diagrams}
    \end{subfigure}
    ~
    \begin{subfigure}[t]{0.32\textwidth}
        \centering
        \includegraphics[width=0.45\columnwidth]{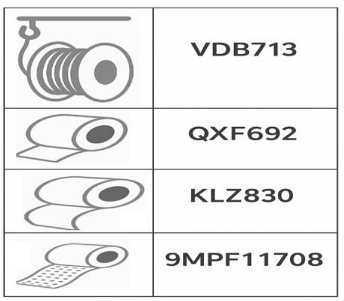}
        \caption{Schematic Diagram}
    \end{subfigure}%
    ~ 
    \begin{subfigure}[t]{0.32\textwidth}
        \centering
        \includegraphics[width=0.55\columnwidth]{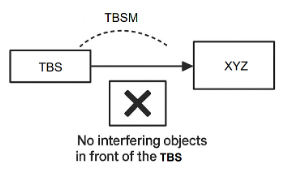}
        \caption{Representation Diagrams}
    \end{subfigure}

    \begin{subfigure}[t]{0.5\textwidth}
        \centering
        \includegraphics[width=0.65\columnwidth]{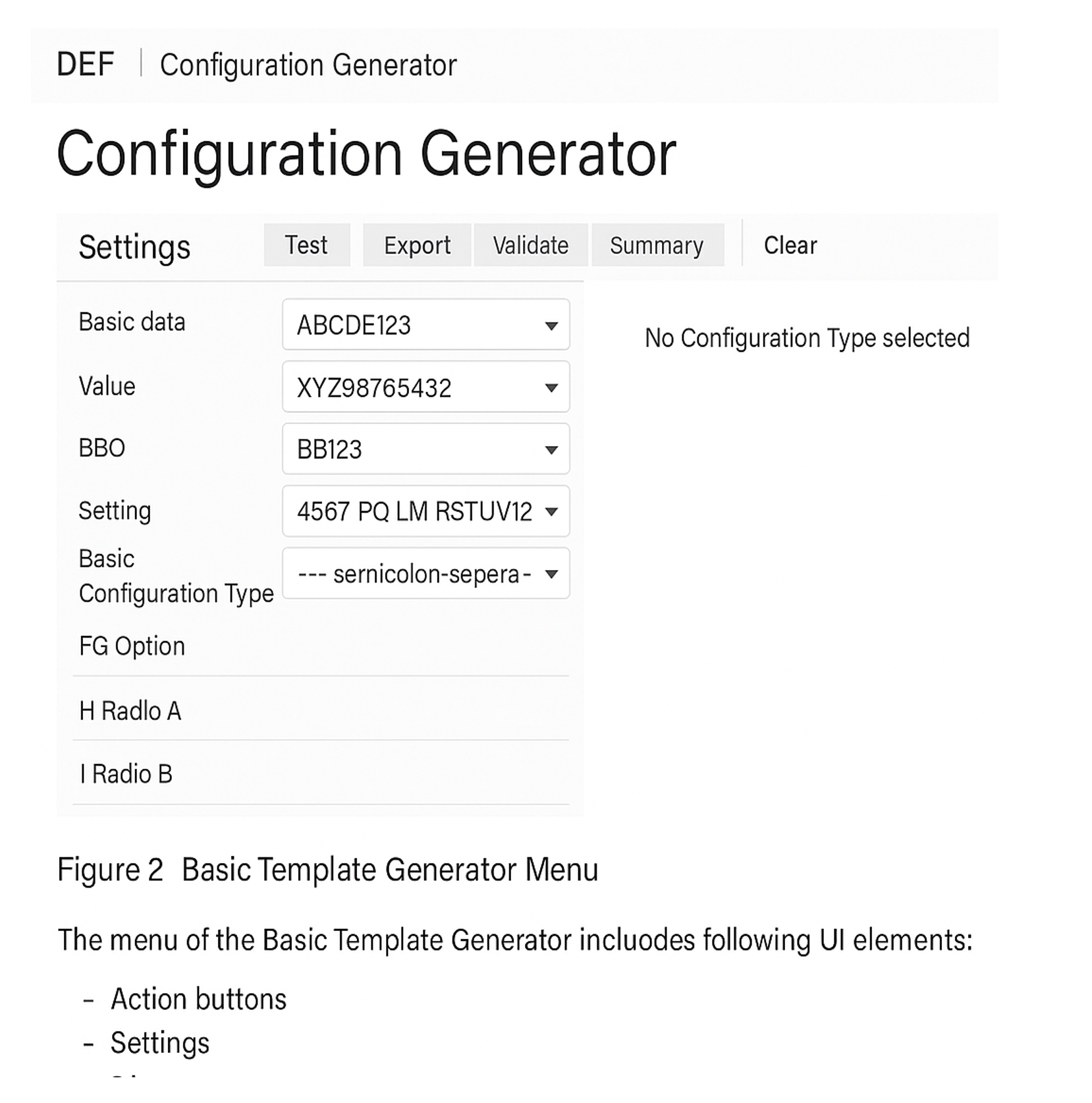}
        \caption{Screenshots}
    \end{subfigure}%
    ~ 
    \begin{subfigure}[t]{0.5\textwidth}
        \centering
        \includegraphics[width=0.75\columnwidth]{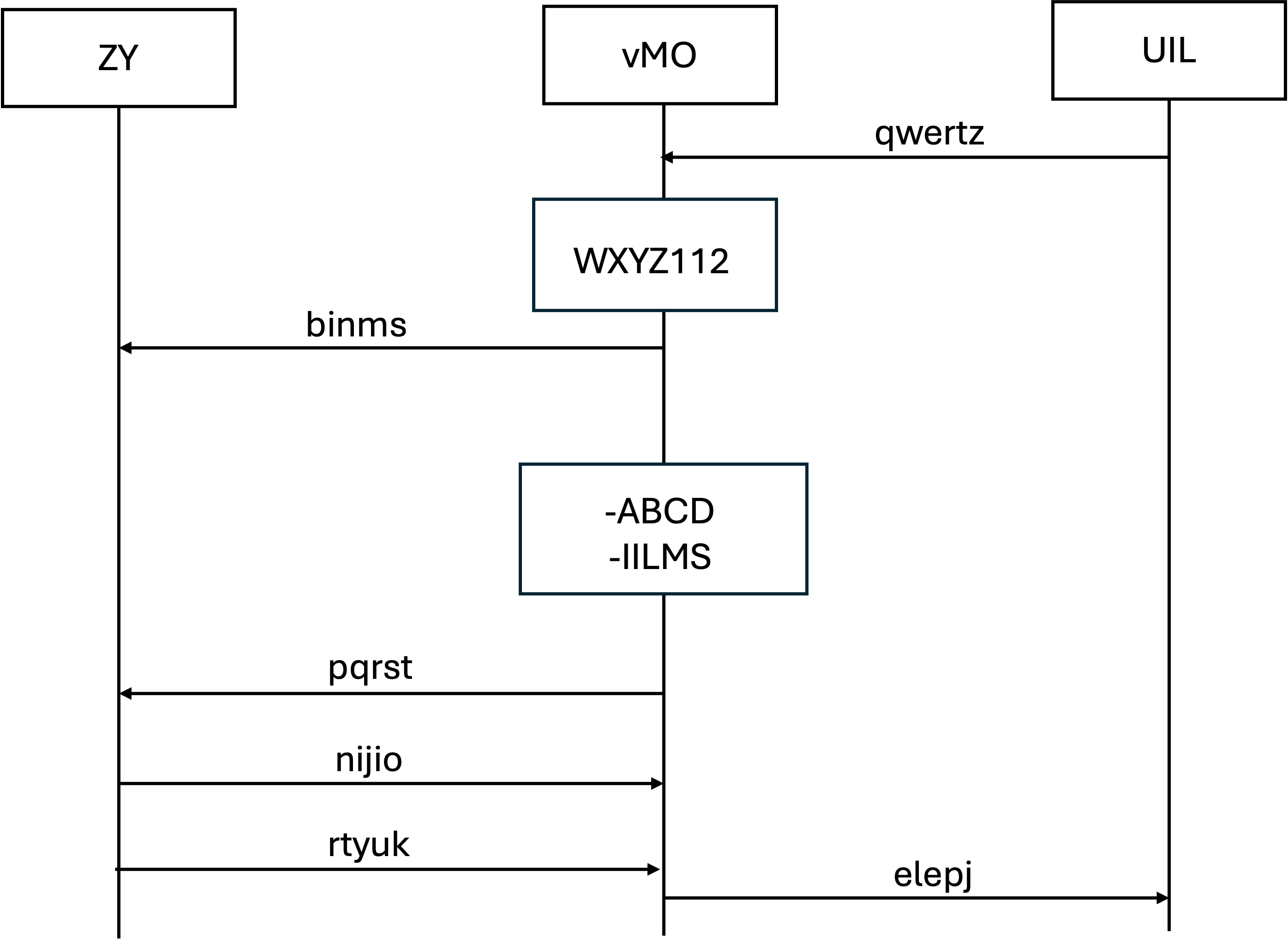}
        \caption{Sequence Diagrams}
    \end{subfigure}
    \caption{Representative images for various categories from the proprietary telecom dataset. (Note: content has been obfuscated based on the confidentiality of data involved.)}
    \label{fig:app_img_categories}
\end{figure*}

Representative examples of various types of images present in the telecom  (PI) dataset are shown in Fig. \ref{fig:app_img_categories}. 

\section{Ambiguous Flowchart images in 3GPP documents}
\label{apdx:3gpp_flowchart}

\begin{figure*}
    \centering
    \includegraphics[width=\linewidth]{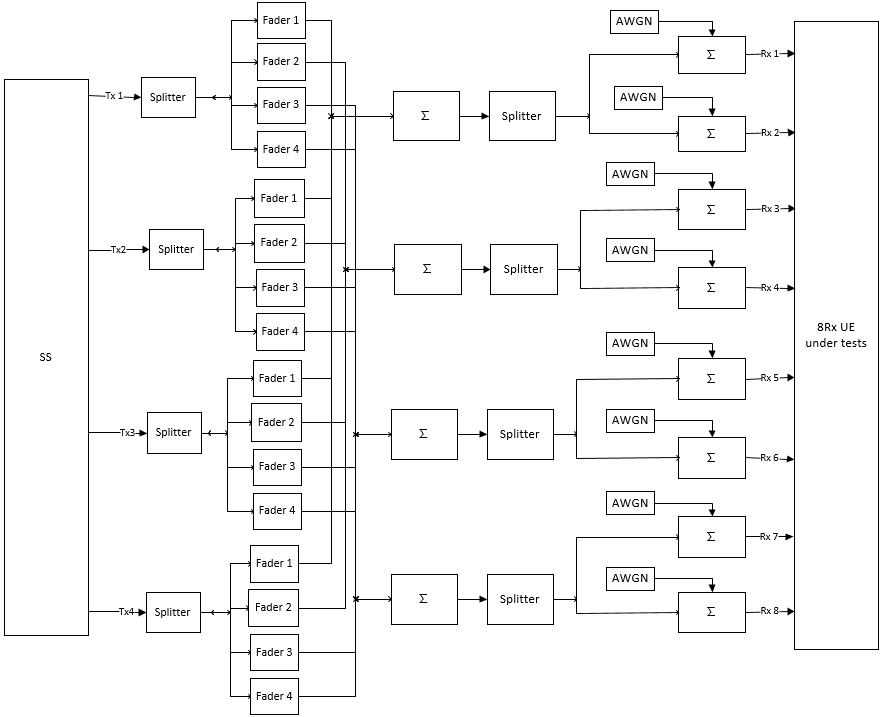}
    \caption{A sample of ambiguous flowchart image in 3GPP document. One can consider this to also be a block diagram.}
    \label{fig:3gpp_flowchart}
\end{figure*}

A representative image from the 3GPP document that is ambiguous as a flowchart is shown in Fig. \ref{fig:3gpp_flowchart}.
\end{document}